\pdfoutput=1

\documentclass[10pt,twocolumn,letterpaper]{article}

\usepackage{cvpr}
\usepackage{times}
\usepackage{epsfig}
\usepackage{graphicx}
\usepackage{amsmath}
\usepackage{amssymb}
\usepackage{listings}
\usepackage{booktabs}

\usepackage[breaklinks=true,bookmarks=false]{hyperref}

\cvprfinalcopy 

\def\cvprPaperID{****} 

\begin{document}

\title{From Local to Global: Navigating Linquistic Diversity in the African Context
}


\author{Nelson Ndugu \thanks{Makerere University, Kampala, Uganda}\\
{\tt\small nelson.ndugu@mak.ac.ug}
\and
Rashmi Margani $^{*}$\\
{\tt\small rashmi.margani@gmail.com}\\
}
%

\maketitle
\begin{abstract} 

This paper focuses on critical problems in NLP related to the linguistic diversity and variation across the African continent, specifically with regards to African local dialects and Arabic dialects that have received little attention. The proposed approach features combination of several word embedding techniques with LightGBM to address the challenge of accurately capturing the context expressed in TUNIZI~\texttt{TUNIZI}. We evaluated our various approaches, demonstrating its effectiveness while highlighting the potential impact of the proposed approach on businesses seeking to improve customer experience and product development in African local dialects.The idea of using the model as a teaching tool for product-based instruction is interesting, as it could potentially stimulate interest in learners and trigger techno entrepreneurship.. Overall, our modified approach offers a promising analysis to the challenges of dealing in African local dialects. Particularly Arabic dialects, which could have a significant impact on businesses seeking to improve customer experience and product development.

\end{abstract}

\pdfoutput=1

\section{Introduction} \label{sect1}

In recent years, social media platforms have become a ubiquitous mode of communication in Africa, generating an enormous amount of data on a daily basis \cite{Barbier2011,Duffett,AdrianWong}. Much of this data includes user-generated content, such as status updates, comments, and reviews, which can provide valuable insights into consumer behavior and sentiment towards products or services \cite{LEE2012331,LITVIN2008458,Muntinga}. However, the analysis of such data can be challenging, particularly when it involves African local dialects that are not widely spoken or have unique characteristics \cite{Olagunju,nekoto2020participatory,adebara2022afrocentric}.
By employing simple sentiment analysis as a starting point through Natural Language Processing techniques, we can overcome linguistic complexity barriers \cite{ogueji-etal-2021-small}, which in turn can lead to the impactful development of AI products in e-commerce, social media networks, banking, healthcare, and other sectors. This will bring diversity and accessibility across the world, especially in areas that have received little attention in NLP inventories for African local dialect data \cite{Olagunju}.
This paper focuses on sentiment analysis of African local dialects, with a particular emphasis on the challenges of analysing text-based data from social media platforms. We propose a novel approach using NLP to extract meaning from various African local dialect communications, highlighting the effectiveness of sentiment analysis in capturing consumers' reviews from social media platforms texts. Our approach demonstrates the potential of AI to help businesses and organisations improve customer experience and product development in the African market.

This paper is organized as follows. In section~\ref{sect2}, we highlight key NLP inventories within African context putting them in business perspectives. In section~\ref{sect3}, we introduce the data and computational models used, following several word embedding techniques and~\texttt{LightGBM} as a classifier. We additionally show the experimental setup adopted and the associated impact of using different balancing technique on the original data sets.  Section~\ref{sect4} and section~\ref{sect5} points the constructed model's applications and the challenges faced, respectively. Section~\ref{sect6} summarizes the model, putting them into perspectives for future investigations.

\section{Problem Context} \label{sect2}

The African continent boasts over 2,000 distinct languages and dialects, some of which have emerged from cultural integrations with foreign cultures \cite{pub1086596433,inbook,beck2010urban}. The Tunisian Arabic(hereafter, TUNIZI) and Kiswahili dialects, for example, are widely spoken in North and East Africa, respectively, and are a fusion of Arabic, French and Latin \cite{Fourati2020TUNIZIAT}; Bantu of coastal East Africa, and Arab traders \cite{Nurse1993SwahiliAS}. These emergent languages are unique and spoken by significant African populations \cite{Fourati2020TUNIZIAT,Nurse1993SwahiliAS}. However, the linguistic diversity of African languages poses a significant challenge for natural language processing, particularly in areas where there is limited knowledge of the indigenous data set \cite{adebara-abdul-mageed-2022-towards}.

In addition to linguistic complexity, word embedding techniques which a very preprocessing step rely on large data sets to generate meaningful vector representations \cite{mikolov2013efficient,pennington-etal-2014-glove,bojanowski-etal-2017-enriching}. Still, the limited availability of data sets for African dialects makes it challenging to create accurate word representations. Furthermore, pre-trained models are often trained on unrelated linguistic data sets and may not accurately capture the nuances and complexities of African languages \cite{Mafunda_Schuld_Durrheim_Mazibuko_2022}. For instance, because certain word embedding techniques may not perform adequately for langauges with typological complexities \cite{chai_2022}. It is possible that some African languages with complex tone systems and those having unique sentence structures unconveying different meanings based on the tone used, while others may have unique sentence structures may not perform well on word embedding algorithms that they were not trained on. To address these challenges, it is crucial to develop word embedding algorithms tailored specifically to African languages and dialects. Because studies show that different word embedding techniques perform differently on different dialects \cite{chai_2022}, we explore the most appropriate word embedding technique for the TUNIZI data set used in this paper in sub section~\ref{expt}.

However, developing such algorithms requires overcoming several challenges. Firstly, limited data sets for African dialects make it difficult to generate accurate representations of words. Furthermore, human errors arising from manual annotations can further degrade the accuracy of word embeddings. The lack of specialized expertise in African languages also poses a significant challenge to the development of natural language processing models for these dialects.

In contemporary times, a significant number of Africans prefer to express themselves on social media using their local dialects or a mix of their dialects with foreign languages such as French or English--a term commonly known as code-switch \cite{sutrisno2021beyond}. However, the challenges discussed above, including the lack of suitable NLP tools, make it difficult for third-party entities such as banking sectors \cite{okolo2021influence}, insurance companies \cite{wang2020customer}, and social media influencers \cite{su15032744socialmediainfluencers} to derive intelligent insights from customers' reviews, likes, and dislikes concerning their products or services \cite{bilgihan2016unified,palmer2020social}. Despite these complexities, This paper demonstrates how capturing customer reviews for African local linguistic dialects can revolutionize the banking, insurance, healthcare, social media, and other sectors

\section{Data and Methods} \label{sect3}

\subsection{Data}
This study utilizes written text messages in the \texttt{TUNIZI} dialect obtained from social media platforms as its primary data source \cite{Chayma2020}. The data was collected using the iCompass platform, which gathered comments expressing sentiment about popular topics. To ensure the data's reliability, it was preprocessed by removing links, emoji symbols, and punctuations. This emergent data was then augmented with the most recent data collected from social media platforms using a web scraper, and further preprocessed using standard NLP techniques, including tokenization, stop word removal, and stemming.

\begin{figure}[ht]
  \centering
  \includegraphics[width=0.5\textwidth]{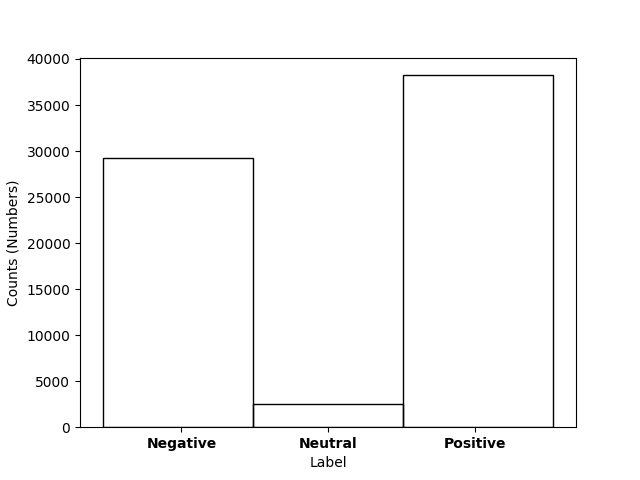}
  \caption{Proportion of the annotated data sets. The positive sentiments are proportionately more than the sentiments for negative and neutral labels (in decreasing order), clearly pointing to imbalancement of the herein dataset.}
  \label{fig:data_hist}
\end{figure}

It is important to note that the \texttt{TUNIZI} dialect represents the Tunisian dialect written in Latin characters and numbers, rather than Arabic letters. As the quality of data directly impacts the performance of machine learning models \cite{Batista2004_balancing_techniques}, we performed a range of diagnostic measures on the preprocessed data, including checking for data imbalances. Our analysis revealed a significant imbalance in the emergent data (see Figure~\ref{fig:data_hist}), which we addressed by balancing the data. 

Imbalanced data can significantly impact the performance of machine learning models by causing them to be biased towards the majority class \cite{chawla2002smote,Haibo2009}. To mitigate this issue, there exist different techniques for handling imbalanced data, each with its strengths and weaknesses. Some of the most commonly used techniques include random undersampling, random oversampling, SMOTE (Synthetic Minority Over-sampling Technique) \cite{chawla2002smote}, ADASYN (Adaptive Synthetic Sampling) \cite{HaiboHe2008}, and cost-sensitive learning \cite{drummond2003c4.5}. We detailed the impact of these techniques on the performance of the emergent model in section~\ref{expt}.

\subsection{Computational Linguistic Model}
Our approach for sentiment analysis utilizes natural language processing (NLP) and involves a combination of two machine learning models, i.e, the  Facebook’s \texttt{FastText}  \cite{bojanowski-etal-2017-enriching,joulin-etal-2017-bag,grave-etal-2018-learning}\footnote{We remind the reader that at later stage several word emdedding technques were considered and their impact on the performance of model are shown in Table~2.} and \texttt{LightGBM} \cite{NIPS2017_6449f44alightgbm}. \texttt{FastText} is used for text feature extraction, and \texttt{LightGBM} is used for sentiment classification. \texttt{FastText} is a lightweight and efficient NLP library that can be used to train text classification models quickly. \texttt{LightGBM}, a gradient boosting framework is optimized for high performance and handles large-scale data efficiently.

Let  $\mathcal{D}={(x_i, y_i)}_{i=1}^{n}$ denote the training dataset, where $x_i$ is the $i$-th text message in the \texttt{TUNIZI} dialect and $y_i \in {-1,0,1}$ is its sentiment label (-1 for negative, 0 for neutral and 1 for positive). Let $f_{\text{fastText}}(\cdot)$ be the function that maps a text message to a fixed-length vector representation using Facebook's fastText algorithm, and let $f_{\text{\texttt{LightGBM}}}(\cdot)$ be the function that maps the vector representation to a sentiment prediction using \texttt{LightGBM}. Then, our sentiment analysis approach can be formulated as follows:

\begin{equation}
f_{\text{sentiment}}(\cdot) = f_{\text{\texttt{LightGBM}}}(f_{\text{fastText}}(\cdot))
\end{equation}

The goal of the approach is to learn the optimal parameters of $f_{\text{fastText}}(\cdot)$ and $f_{\text{\texttt{LightGBM}}}(\cdot)$ on the training dataset $\mathcal{D}$, so that the resulting $f_{\text{sentiment}}(\cdot)$ can accurately predict the sentiment of new text messages in the \texttt{TUNIZI} dialect. This can be achieved by minimizing the following loss function \cite{broyden1970nonlinear}:

\begin{equation}
\mathcal{L}(\theta) = \frac{1}{N} \sum_{i=1}^{N} \ell(y_i, f_{\text{sentiment}}(x_i;\theta)),
\end{equation}

where $\theta$ denotes the parameters of $f_{\text{fastText}}(\cdot)$ and $f_{\text{\texttt{LightGBM}}}(\cdot)$, $\ell(\cdot)$ is the binary cross-entropy loss function, and $f_{\text{sentiment}}(x_i;\theta)$ is the predicted sentiment of $x_i$ given the parameters $\theta$. The optimal parameters $\theta^*$ can be found by solving the following optimization problem:

\begin{equation}
\theta^* = \arg\min_{\theta} \mathcal{L}(\theta)
\end{equation}

Once the optimal parameters are obtained, the sentiment of a new text message $x$ can be predicted as follows:

\begin{equation}
\hat{y} = f_{\text{sentiment}}(x;\theta^*)
\end{equation}

where $\hat{y}$ is the predicted probability for the sentiment label (-1 for negative, 0 for neutral and 1 for positive). The multi-polarity classification requires a threshold probability parameter, $t$.  If $\hat{y}\geq t$, the text is classified as positive; if $\hat{y} \leq 1-t$, it is classified as negative; and if $1-t < \hat{y} < t$, it is classified as neutral. The choice of threshold $t$ can affect the performance of the predictive model. A lower threshold will classify more texts as positive, while a higher threshold will classify more texts as negative. The optimal threshold can be found by selecting the value that maximizes the f1 score--which is a measure of the balance between precision and recall \cite{abbasi-etal-2014-benchmarking}. In this model there was no threshold specification, the LightGBM will therefore assign the predicted class label based on the class label with the highest probability. For example, if the predicted probabilities for a given input text are [0.2, 0.3, 0.5] for multi-class label [0, -1, 1], LightGBM will assign the class label 1, which has the highest probability.

\subsection{Experimental Setup}\label{expt}
We implemented our approach using Python and the scikit-learn and LightGBM libraries. The models were trained and evaluated on a machine with an Intel Core i7 processor, 16GB of RAM, and an NVIDIA GeForce GTX 1050Ti GPU. We used a 70-30 split for training and testing the models, respectively. The hyperparameters of the models were tuned using grid search with 5-fold cross-validation. Our model was evaluated using f1-score.

In the case of African local dialect linguistic sentiment analysis, the dataset may be imbalanced due to differences in the frequency of certain sentiments in the language. For example, some sentiments may be expressed more frequently than others, leading to an imbalanced dataset. The f1-score takes into account both precision and recall, which are important measures in sentiment analysis. Precision measures the proportion of true positive classifications out of all the positive classifications made by the model, while recall measures the proportion of true positive classifications out of all the actual positive samples in the dataset.
In sentiment analysis, a high precision means that the model is correctly identifying positive or negative sentiment when it is present, while a high recall means that the model is correctly identifying all positive or negative sentiment in the dataset.
Therefore, by using the f1-score, which balances both precision and recall \cite{sasaki2007truth, boughorbel2017optimal}, we can get a better overall measure of the performance of the sentiment analysis model in African local dialect linguistics. It helps in measuring how well the model is performing across all classes, and not just the majority class, which is important when dealing with imbalanced datasets.

Parametric sweeps were performed on the nominal model by executing five folds across 50 parametric combinations, giving a total of 250 fits. After this operations, we achieved a best f1-score of 0.8065 for the nominal model with the following model parameters:

\begin{verbatim}
{'subsample': 0.6, 'reg_lambda': 1.0, 
'reg_alpha': 0.1, 'num_leaves': 64, 
'n_estimators': 1000, 'min_child_weight': 10,
'max_depth': 8, 'learning_rate': 0.1, 
'colsample_bytree': 1.0}
\end{verbatim}

Because NLP models are sensitive to the training data, and balancing contributes to data set diversities, we test how different balancing techniques on the original imbalanced data affect the performance of the  model. The performance from different balancing techniques, in particular, \texttt{Under-Sampling},~\texttt{Over-Sampling} and~\texttt{ADASYN} were compared against the nominal model without any balancing.  To evaluate the performance of the models against different balancing techniques, hence dataset diversities, we used f1-scores, and the results are summarized in Table~\ref{tab:imbalanceperform}.
\begin{table}[ht]
\small
  \centering
  \caption{Sensitivity of our model to the different data balancing techniques. \texttt{US},~\texttt{OS} and~\texttt{ADASYN} denote under sampling, over sampling and adaptive synthentic sampling techniques, respectively.}
   \label{tab:imbalanceperform}
  \begin{tabular}{@{}cccccc@{}}
    \toprule
    Technique & Learning rate& Max depth& \# Estimators & f1-score \\
    \midrule
    \texttt{US} & 0.01 & 6 & 1000 & 0.7036 \\
    ~\texttt{OS} & 0.1 & 8 & 1000 & {\bf 0.9071} \\
    ~\texttt{ADASYN} & 0.1 & 8 & 1000 & 0.8640 \\
    \bottomrule
  \end{tabular}
\end{table}

This analysis aimed to determine the most effective balancing technique to use with our NLP model. The results summarized in the table suggest that Over-Sampling is the most effective technique for addressing the class imbalance issue and improving the performance of the model on all classes. Over-Sampling had the highest f1-score of {\bf 0.9071}, which indicates that the model was able to better capture the nuances and complexities of the dataset and produce more accurate results.

The results also show that ADASYN, which creates synthetic samples to address class imbalance, performed better than Under-Sampling, which simply removes instances from the majority class. This suggests that creating synthetic samples can be an effective technique for addressing class imbalance in the TUNIZI dataset and improving model performance.

Furthermore, the hyperparameters selected for each technique also played a role in the results obtained. A higher learning rate and greater depth can allow the model to better capture the complexity of the dataset, while increasing the number of estimators can improve the model's accuracy. However, the optimal hyperparameters can vary depending on the dataset and the specific problem being addressed.

Overall, the results suggest that careful selection of appropriate techniques and hyperparameters is critical for achieving good model performance. Especially when little is known about the appropriate word embedding techniques and the model is trained on complex and imbalanced dataset, like the TUNIZI data set. The findings of this study are essential in guiding practitioners towards the adoption of the most appropriate balancing technique for enhancing the performance of NLP models within imbalanced African local dialect dataset

Besides ~\texttt{fasttext} (herein nominal word embedding technique), the different word embedding techniques considered in this study are~\texttt{Contextual Word Embedding} and~\texttt{Byte Pair Encoding (BPE)}. The ~\texttt{fasttext},~\texttt{BERT} and~\texttt{Byte Pair Encoding (BPE)} word embedding techniques, feature count-based methods.  The results of our analysis are presented in Table~\ref{tab:embperform} below.

\begin{table}[ht]
\small
  \centering
  \caption{Influence of the different word embedding techniques on the model performance. CWE denotes Contextual Word Embedding.}
   \label{tab:embperform}
  \begin{tabular}{cccccc}
    \toprule
    Model & Learning rate& Max depth& \# Estimators & f1-score \\
    \midrule
    \texttt{fasttext} & 0.1 & 8 & 1000 & {\bf 0.8065} \\
    ~\texttt{CWE} & 0.1 & 8 & 1000 & 0.7019 \\
    ~\texttt{BPE} & 0.05 & 8 & 500 & 0.6944 \\
    \bottomrule
  \end{tabular}
\end{table}

The table shows the results of three different word embedding techniques - FastText, Contextual Word Embedding, and BPE - on the TUNIZI dataset.

Each of these techniques was trained using different hyperparameters, which are shown in the table. Specifically, the learning rate, maximum depth, and number of estimators were varied across the different models.

The results show that FastText performed the best overall, with an f1-score of 0.8065. This suggests that FastText was able to capture more of the nuances and complexities of the TUNIZI dataset compared to the other models.

Contextual Word Embedding and BPE both performed worse than FastText, with f1-scores of 0.7019 and 0.6944, because contextual word embedding models like BERT and Byte pair embedding models like GPT-2, RoBERTa, BART, and DeBERTa impose many challenges such as:
\begin{itemize}
 \item [(1)] Lack of data: The training data for these models is predominantly in English, which leads to poor performance for African languages due to limited availability of training data.
 \item [(2)] Limited vocabulary: Many African languages have a complex morphology with rich inflectional and derivational aspects, resulting in a vast vocabulary. However, these models have a limited vocabulary size, which makes it challenging to represent the full range of meaning in African languages.
 \item [(3)] Linguistic diversity: African languages are diverse and have different dialects, accents, and writing systems, which makes it challenging to train a model that can handle this variation. In contrast, FastText uses sub-word units to represent words, which makes it more effective for languages with complex morphology and limited training data. FastText can also handle linguistic diversity better than BERT and other models because it does not rely on pre-existing language models. Instead, it learns embeddings from the training data directly.
\end{itemize}

Overall, these results highlight the importance of selecting appropriate hyperparameters, experimenting with different word embedding techniques and  models as well as the need to evaluate the performance of these models using appropriate metrics such as the f1-score.

\section{Application of our Model} \label{sect4}

Measuring consumers' reviews from social media platforms can be achieved using the developed model through sentiment analysis. Sentiment analysis involves analyzing the polarity of a given text, which in our case was measured as either positive, neutral, or negative . The developed model can be used to classify texts into these categories. Suppose a company wants to measure consumers' reviews from social media platforms regarding a product they recently rolled out. They can deploy the developed model to experiment with the sentiment analysis.

Given a new social media text, $x$, the model outputs a predicted probability, $\hat{y}$, which represents the probability of the text belonging to a particular sentiment category. The predicted probability, $\hat{y}$, can then be used to classify the text into one of the three categories based on a threshold, $t$.

Companies can use this sentiment analysis to monitor the consumers' reviews regarding their product and make necessary improvements based on the feedback. They can also use the sentiment analysis to compare their product's sentiment with their competitors' products and make informed decisions to stay ahead of the competition. The developed model can be a powerful tool for businesses to measure consumers' reviews from social media platforms and other sectors to make data driven decisions.

\section{Challenges} \label{sect5}
In this paper, we developed a model for sentiment analysis and applied it to measure consumers' reviews from social media platforms. However, the process of developing such models is not without challenges. In this section, we discuss some of the challenges we encountered during the development of our model:

\begin{itemize}
 \item [(1)] Obtaining high-quality training data; The first challenge we encountered was obtaining high-quality training data. In order to train our model, we needed a large corpus of social media texts that were manually annotated with the corresponding sentiment. This was a time-consuming and labor-intensive process. Furthermore, we had issues with the reliability and consistency of the annotators, which could have impacted the accuracy of the final model. To mitigate this challenge, we employed various strategies such as using multiple annotators and implementing quality control checks to ensure the accuracy and consistency of the annotations.
 \item [(2)] Dealing with the inherent ambiguity and complexity of natural language was a significant challenge in our project, especially in the Tunisian context. Tunisians tend to use informal language in social media platforms, which adds an extra layer of complexity to the task of sentiment analysis.

Different people in Tunisia may use the same words or phrases to express different reviews depending on their cultural and social backgrounds. Moreover, context plays a crucial role in determining the sentiment of a text. For example, the word "hlib" in Tunisian Arabic can mean "bread," but it can also be used as a slang term to refer to money. Therefore, it is essential to consider the context in which a word is used to understand its sentiment properly.

To illustrate this point, consider the following two sentences in Tunisian Arabic:

\begin{itemize}
\item [(a)] "Lebnen mte3i tayyeb, mais el khedma makanch tayba."
\item [(b)] "Lebnen mte3i mchayba, mais el khedma tayyba."
\end{itemize}

Both sentences contain positive and negative words, but the overall sentiment of the sentences is different. The first sentence means "My sandwich was good, but the service was bad," while the second sentence means "My sandwich was bad, but the service was good." This highlights the importance of considering the context and the overall meaning of a text to determine its sentiment accurately.

To overcome these challenges, we employed careful feature engineering and model selection techniques to ensure that our model could capture the nuances of Tunisian Arabic and make accurate predictions. We also used a large corpus of annotated data to train and test our model, which helped to improve its accuracy.

Overall the inherent ambiguity and complexity of natural language was a significant challenge in our project, especially in the Tunisian context. However, by employing careful feature engineering and model selection techniques and using a large corpus of annotated data, we were able to develop a model that could accurately predict the sentiment of social media texts in Tunisian Arabic.

\item [(3)] Technical challenges related to implementation and deployment: Finally, there were technical challenges related to the implementation and deployment of the model. The large size of the training data and the complexity of the model required significant computational resources, and there were issues with optimizing the model for efficiency and scalarbility. Additionally, there were challenges related to integrating the model with existing software systems and deploying it to production environments. These challenges required significant expertise in machine learning, software engineering, and DevOps, and posed a significant barrier to adoption for organizations without these resources.
\end{itemize}

In summary, developing a model for sentiment analysis is a challenging task for African local dialect data requiring careful consideration of various factors such as data quality, model complexity, and deployment requirements. By addressing the challenges discussed in this paper, we were able to develop a model that could accurately classify social media texts into positive, negative, or neutral categories. However, it is important to note that the challenges we faced are not unique and are likely to be encountered by others who attempt to develop similar models. Therefore, it is important to be aware of these challenges and develop appropriate strategies to overcome them.
\section{Conclusions}
\label{sec:conclusions} \label{sect6}
Our paper provides a comprehensive analysis and guide on how to approach NLP problems within the TUNIZI and other African dialect dataset. This has the potential to revolutionize the African business market and enhance transparency and diversity as it enables the use of African local dialect languages in communication with the rest of the world through language translation, chat conversation, etc. Leveraging established NLP techniques, we augmented the TUNIZI data with current text from social media platforms, resulting in a comprehensive dataset. Our model, embedded using Facebook's FastText and classified using the LightGBM classifier we developed, achieved an impressive f1-score of 0.8065, after a thorough parametric experimentation.

Despite the inherent complexities of the TUNIZI dataset, our model shows promise in predicting customers' reviews with high accuracy. We believe that our approach could generate data-driven insights for decision-making during product development, empowering businesses to meet the needs and expectations of their TUNIZI-speaking customers more effectively. Moreover, with further training on various African dialects, our model could predict customer reviews not only for TUNIZI dialects but also for several other African dialects.

This paper suggests three future perspectives to further improve the sentiment analysis models within the diverse African dialect context:
\begin{itemize}
 \item [(1)] Clustering the diverse African languages based on dialect similarities and create specific word embedding techniques for each cluster. 
 \item [(2)] Gathering diverse African dialects with annotation guided by local dialect experts.
  \item [(3)] Experimenting with different word embedding techniques and using the one that optimizes the model's performance in case of a lack of knowledge of the appropriate word embedding technique.
\end{itemize}

We believe this moderate level innovation has the potential of positively influencing the education sectors. For example, it has the capability of inspiring product-based teaching methods, where STEM learners can derive motivation from the prowess of our model and use it to spark innovation in their communities.
\section*{Acknowledgements}
The work in this paper was partially supported by KaggleX BIPOC mentorship program.

{\small
\bibliographystyle{ieee}
\bibliography{biblio}
}

\end{document}